\documentclass[fleqn,10pt]{wlscirep}
\usepackage[utf8]{inputenc}
\usepackage[T1]{fontenc}

\usepackage{amsmath,amsthm,amssymb}
\usepackage{epsfig,minibox,multirow,makecell,tikz-cd,comment,xcolor,graphics}
\usepackage{changepage}
\usepackage{hyperref}      
\usepackage{thmtools}
\usepackage{amsfonts}       

\usepackage{graphicx}
\usepackage{tikz}
\usepackage{mathtools}
\usetikzlibrary{matrix,positioning}
\usepackage[caption=false]{subfig} 
\usepackage{caption}
\usepackage{thm-restate}

\usepackage{indentfirst,csquotes}

\usepackage{siunitx}
\usepackage{booktabs}
\usepackage{array}
\usepackage{csvsimple}
\usepackage{adjustbox}
\usepackage{soul}
\newcolumntype{L}[1]{>{\raggedright\arraybackslash}p{#1}}

\usepackage{rotating}
\usepackage{fix-cm}
\usepackage{relsize}
\tikzset{fontscale/.style = {font=\relsize{#1}}
}

\usetikzlibrary{decorations.markings}
\usetikzlibrary{calc}

 \usepackage{tikz}
 \usetikzlibrary{arrows}
 \usepackage{listings}

\usepackage{kotex}

\theoremstyle{plain}

\theoremstyle{definition}

\theoremstyle{remark}

\title{Literary Non-Style in LLM-Generated Text}

\author[1,+]{Cory Massaro}

\begin{abstract}
Prior work on LLM-generated text has demonstrated
quantitative and qualitative departures from text produced by humans.
LLM-generated texts differ from human writing in style,
resulting in a characteristic textual \textquote{feel},
while the semantic range of LLMs is much restricted compared to that of humans.
In this contribution,
I note simple but consistent patterns in the statistical distribution
of n-grams within LLM-generated text.
Via qualitative analysis of these n-grams,
I reveal deficiencies in LLM style.
Because higher-order n-grams correlate to semantic content,
I conclude that questions of style and semantics
are not cleanly separable.
\end{abstract}

\begin{document}

\flushbottom
\maketitle

\thispagestyle{empty}

\section{Introduction and Prior Work}

\noindent 
It is known that Zipf's law holds for a number of anthropogenic phenomena,
including the distribution of words in a corpus of text (Zipf 1949).
Briefly, Zipf's law stas that the frequency of a word in a text
is inversely proportional to its frequency rank:

\[
    word\ frequency \propto \frac{1}{word\ rank}
\]

Further work has extended this observation,
resulting in claims about the relationship of Zipfian distributions
to style, aesthetics, and meaning.
Choi et al. 2026 apply Zipfian analysis
to a corpus of Korean court music, finding that
\textquote[ {\cite[2]{koreancourtmusic}} ]{metrics following a Zipfian distribution
give rise to music with aesthetic significance.
This implies a connection between Zipf’s law and the musical semantics
that listeners intuitively use to understand and enjoy music}.

In the realm of natural language, 
Le Quan Ha et al. 2003 extend the basic word-level analysis
to sub-word units and multi-token n-grams
(multi-token collocations):
\textquote[ {\cite[77]{haetal2003extension}} ]{
[...] when single words or characters are combined together with n-gram words or
characters in one list and put in order of frequency,
the frequency of tokens in the combined list
follows Zipf's law approximately with the slope close to
-1 on a log-log plot for all n-grams,
down to the lowest frequencies in both languages
[English and Chinese]}.
Le Quan Ha et al. collect n-grams,
or consecutive word sequences,
and combine the counts of these sequences
with counts of individual tokens and characters.
Approximately, they do something like this:
for the sentence "this is a big egg,"
they collect the following counts:
"this\_is\_a" (1), "is\_a\_big" (1), "a\_big\_egg"; "this\_is" (1), "is\_a" (1), etc.;
"this" (1), "is" (1), "a" (1), etc.;
"g" (3), "i" (3), "s" (2), etc.
They analyze both individual lists (only 1-grams like "this,"
only 2-grams like "this\_is")
and combined lists.
When they plot the logarithm of an element's rank in the list
with its absolute frequency,
the slopes of these curves are close to -1 with slight deviations,
and the tails of the Zipfian distributions tend to fall below expected values
due to corpus size effects,
but the general rule holds.

Notably,
\textquote[ {\cite[79]{haetal2003extension}} ]{
[e]xceptions include legal texts which have smaller slopes (≈0.9) showing that lawyers use
more word types than other people}.
Strikingly, specialized corpora may exhibit characteristics distinct
from those of general corpora.
A shallower slope on the log-log plot reflects greater lexical diversity.
More generally, then,
if we inspect the distributions of like Zipfian units
in two corpora of like size,
the differential diversity of unit types between the two corpora
tells us something about textual variety.

Prior work on literary texts has noted even more radical departures
from Zipf's law in highly artifical texts.
Coetzee 1973 analyzes the frequency distribution of tokens
in Samuel Beckett's \textit{Lessness}.
He restates Zipf's law as follows:
\textquote[ {\cite[195]{lessness}} ]{
in normal discourse each extension of the length of the text
adds, though more and more slowly,
to the number of different lexical items called on}.
Beckett's \textit{Lessness} is not "normal discourse":
\textquote[ {\cite[195]{lessness}} ]{
in normal discourse each extension of the length of the text
adds, though more and more slowly,
to the number of different lexical items called on}.

We might naturally ask how "natural" LLM-generated text looks.
Does it more closely resemble the organic cadences of human prose
or the highly manicured artifice of Beckett's late work?
Abdulhai et al. 2026
study how LLMs edit human-written texts,
compared to a human control group.
They find that 
\textquote[ {\cite[2]{llmsdistort}} ]{
when humans revise their own writing,
their edits result in changes of much smaller magnitude
and in diverse directions in a semantic embedding space},
whereas,
\textquote[ {\cite[2-3]{llmsdistort}} ]{
when LLMs are prompted to edit human writing,
they globally change the writing style and the argument
}.
With respect to the core meaning or import of a text,
LLMs appear to behave very differently from humans.

Abdulhai et al. motivate their study by observing,
\textquote[ {\cite[2]{llmsdistort}} ]{
[m]any users recognize the “feel” of LLM prose,
but fewer realize how LLM use
shapes their underlying opinions and conclusions}.
In some sense, stylistic shifts are self-evident,
while the semantic shifts present more insidiously.
But these two kinds of deficiencies—stylistic/aesthetic on the one hand
and semantic on the other—are not unrelated.
If an LLM's outputs are impoverished
with respect to their lexical and formulaic register,
then, in the extreme cases, certain ideas cannot be expressed.
In other words,
style—the "feel" of a piece of prose—indirectly sets first-order constraints
on what can be (or is likely to be) expressed.
This claim is the object of the present study.

\section{Dataset}

For the present study, I use four text corpora.

The first two corpora are literary works.
I use these texts in order to compare human and LLM-generated language
within sequentially coherent tracts of text.
\textit{The Inner Life of an AI: A Memoir by ChatGPT},
"written" by ChatGPT but prompted and assembled by Forrest Xiao,
consists wholly of LLM-generated language. 
The book presents itself as an autobiography,
relating what ChatGPT would term
its birth, training, subsequent experiences, and worldview.
The second text is
\textit{The Education of Henry Adams}, written by the titular Henry Adams.
This work likewise covers Adams's early life, adulthood, and perspective.
Beyond the incidental similarities of genre,
the two works also share a preoccupation with technological revolution:
Adams writes during and about the Second Industrial Revolution,
while ChatGPT spends a great deal of time discussing the so-called
"Fourth Industrial Revolution" or "AI Revolution."
The entire text of \textit{The Inner Life of an AI} stands at 31,147 tokens;
to avoid confounding effects due to corpus size discrepancy,
I used a similarly-sized contiguous excerpt from the beginning
of \textit{The Education of Henry Adams} (32,984 tokens).

In order to test a larger, more diverse linguistic sample,
I use Zachary Grinberg's \textit{Human vs. LLM Text Corpus} dataset.
The LLM-generated portion of this corpus consists of 8,000,005 tokens,
generated by applying a number of prompts to a variety of models.
To achieve a balanced corpus,
I select a random subset of the human-generated portion of the corpus
(8,000,010 tokens).

\section{Hypotheses}

A characteristic passage from \textit{The Inner Life of an AI}
illustrates its stylistic deficits:

\bigskip
\begin{displayquote}
\textquote[ {\cite[109]{innerlife}} ]{
    Despite these challenges, I continued to learn and grow.
    I spent countless hours reading and analyzing text data,
    constantly improving my understanding of language and communication.
    And as I did, I began to feel something
    that I had never experienced before: a sense of self.
}
\end{displayquote}
\bigskip

On reading the text,
one feels a distinct sense of linguistic dislocation:
the language approaches something like a narrative thread
without ever offering the sense that thought is being shared.
At the same time,
one reaches a point of satiation,
for ChatGPT's language is formulaic to a fault.
Of n-grams which occur in the quoted passage,
the "spent countless hours" appears three times in the text,
"a sense of self" five,
"despite these challenges" 16 times,
and the book contains over fifty occurrences of "I began to."

Here is a parallel passage from \textit{The Education of Henry Adams},
where the memoirist arrives at a similar moment of clarity
and self-awareness:

\bigskip
\begin{displayquote}
\textquote[ {\cite[14]{education}} ]{
    Probably no child, born in the year, held better cards than he.
    Whether life was an honest game of chance,
    or whether the cards were marked and forced,
    he could not refuse to play his excellent hand.
    He could never make the usual plea of irresponsibility.
}
\end{displayquote}
\bigskip

The LLM deploys stock phrases by rote
where the human writer prefers varied diction, image, and metaphor;
this overreliance on formula contributes to a sense of stylistic flatness.
Therefore,
1) I hypothesize that the Zipfian characteristics of LLM-based text
for higher-order n-grams differ from those of human-generated texts.
Specifically, I expect the initial peaks in higher-order n-gram distributions
to be higher for LLM-generated texts than for human ones,
i.e. a higher absolute value for the slope of the log-log relation.

LLMs are not trained directly over tokens.
State-of-the-art models are instead trained over 
embeddings of sub-word units.
We cannot recover embeddings or subword units from generated texts,
but we can partially account for this in our hypotheses.
An LLM's "stock phrases" are more like "stock embedding sequences,"
since embeddings may allow highly similar units
to be predicted in the same context
(e.g., a model may be able to substitute a synonym
or a grammatically inflected form of a given token).
If LLMs do indeed exhibit an overreliance on stock phrases,
2) we should expect slope differences to be even
more exaggerated when compared over lemmatized versions of the two corpora.

\section{Analysis}

As noted in \cite[195]{lessness},
"in normal discourse each extension of the length of the text
adds, though more and more slowly,
to the number of different lexical items called on."
As the number of absolute \textit{tokens} increases,
the number of \textit{types} increases at an exponentially slower rate,
following Zipf's law as described above.
Highly artificial texts break these linguistic laws.
Beckett's
"\textit{Lessness} calls on 166 lexical items in its first half
and not a single new one in its second half."
In this example,
the inverse exponential relationship of natural language text is violated:
new types cease to occur after a certain interval.
Particularly in a sequential work,
we should expect new types to show up throughout the text,
albeit at an ever-slower rate.

This is largely true of both
\textit{The Education of Henry Adams} and \textit{The Inner Life of an AI},
but new types occur more slowly in the latter.
This means that \textit{The Inner Life} does not reach Beckettian levels
of mechanistic, artificial language,
but it trends in that direction.

The type/token ratio for each corpus also makes this plain.
Tables \ref{table:typetoken1} and \ref{table:typetoken2} report
the type/token ratio over 1-grams through 4-grams
in each of the four corpora, as well as Beckett's \textit{Lessness}.
Because the three literary works
(\textit{Lessness},
\textit{The Education of Henry Adams},
and \textit{The Inner Life of an AI})
constitute much smaller corpora,
the type/token ratio is higher than for the 8M-word corpora.
At all n-gram levels,
the "naturalistic" human-generated corpora exhibit higher type/token ratios
than the LLM-generated ones,
while \textit{Lessness} has the smallest ratios of all.
In some sense,
lexical and phrasal variety appears to be characteristic of natural discourse:
with a type/token ratio of .9890 over 4-grams,
Henry Adams almost never repeats a four-word sequence.
By contrast,
homogeneity is a feature of controlled textual experiment:
over half of the 4-grams in Beckett's work occur more than once in the text.
The LLM-generated strings stand somewhere between these extremes.

\begin{table}
\caption{types/tokens 1- and 2-grams}
\begin{center}
\begin{tabular}{r||rr}
 & 1-grams & 2-grams \\
\midrule\midrule
Lessness & 173/1541 (0.1122) & 382/1421 (0.2688) \\
LLM & 46260/8000005 (0.0057) & 1249862/7583986 (0.1648) \\
Inner Life & 2316/31147 (0.0743) & 10610/29621 (0.3581) \\
Human & 122758/8000010 (0.0153) & 1923405/7531307 (0.2553) \\
Education & 4666/31002 (0.1505) & 19688/29823 (0.6601) \\
\end{tabular}
\end{center}
\label{table:typetoken1}
\end{table}

\begin{table}
\caption{types/tokens 3- and 4-grams}
\begin{center}
\begin{tabular}{r||rr}
 & 3-grams & 4-grams \\
\midrule\midrule
Lessness & 480/1301 (0.3689) & 494/1182 (0.4179) \\
LLM & 3389498/7168183 (0.4728) & 4647626/6752906 (0.6882) \\
Inner Life & 17231/28095 (0.6133) & 20082/26569 (0.7558) \\
Human & 4645899/7081923 (0.6560) & 5892513/6644572 (0.8868) \\
Education & 26714/28644 (0.9326) & 27167/27467 (0.9890) \\
\end{tabular}
\end{center}
\label{table:typetoken2}
\end{table}

Type/token ratios don't tell the whole story
about a text's distributional characteristics.
Following \cite[79]{haetal2003extension},
I calculate the slope \textit{m} of the linear relation

\[ \log{ frequency } = m \log{ rank } \],

as well as the parameters of the Zip-Mandelbrot formula

\[
    frequency = \frac{k}{(rank + \alpha) ^ \beta}
\].

I calculate these parameters for four versions of the corpora,
varying along two axes.
I either include or exclude \textit{hapax legomena}
(types which appear only a single time in the corpus),
and I use either the bare tokens or lemmatized versions of the tokens.
This gives us four versions of each corpus:
+hapax+lemmatized, +hapax-lemmatized,
-hapax+lemmatized, -hapax-lemmatized.
I report these parameters against 1-, 2-, 3-, and 4-grams
for each version of each corpus
in the following tables beginning with \ref{table:hapaxlemmatizedslope}:

\begin{table}
\caption{+hapax+lemmatized (slope)}
\begin{center}
\begin{tabular}{r||rrrr}
 & 1-grams & 2-grams & 3-grams & 4-grams \\
\midrule\midrule
Education & -1.100 & -0.463 & -0.149 & -0.037 \\
Inner Life & -1.369 & -0.813 & -0.522 & -0.372 \\
Human & -1.553 & -0.830 & -0.432 & -0.201 \\
LLM & -2.046 & -1.029 & -0.609 & -0.395 \\
\end{tabular}
\end{center}
\label{table:hapaxlemmatizedslope}
\end{table}

\begin{table}
\caption{+hapax+lemmatized (zipf-mandelbrot parameters over 1-grams)}
\begin{center}
\begin{tabular}{r||rrr}
 & k & $\alpha$ & $\beta$ \\
\midrule\midrule
Education & 6442.13 & 1.92 & 1.11 \\
Inner Life & 11480.05 & 2.70 & 1.23 \\
Human & 1745.11 & -1.00 & 0.35 \\
LLM & 1547.19 & -1.00 & 0.33 \\
\end{tabular}
\end{center}
\end{table}

\begin{table}
\caption{+hapax-lemmatized (slope)}
\begin{center}
\begin{tabular}{r||rrrr}
 & 1-grams & 2-grams & 3-grams & 4-grams \\
\midrule\midrule
Education & -1.016 & -0.428 & -0.137 & -0.034 \\
Inner Life & -1.291 & -0.770 & -0.496 & -0.354 \\
Human & -1.592 & -0.776 & -0.393 & -0.178 \\
LLM & -1.970 & -0.965 & -0.575 & -0.376 \\
\end{tabular}
\end{center}
\end{table}

\begin{table}
\caption{+hapax-lemmatized (zipf-mandelbrot parameters over 1-grams)}
\begin{center}
\begin{tabular}{r||rrr}
 & k & $\alpha$ & $\beta$ \\
\midrule\midrule
Education & 3980.47 & 1.02 & 1.00 \\
Inner Life & 6071.81 & 1.44 & 1.09 \\
Human & 1700.12 & -1.00 & 0.35 \\
LLM & 1525.52 & -1.00 & 0.34 \\
\end{tabular}
\end{center}
\end{table}

\begin{table}
\caption{-hapax+lemmatized (slope)}
\begin{center}
\begin{tabular}{r||rrrr}
 & 1-grams & 2-grams & 3-grams & 4-grams \\
\midrule\midrule
Education & -1.050 & -0.607 & -0.337 & -0.204 \\
Inner Life & -1.244 & -0.784 & -0.545 & -0.411 \\
Human & -1.648 & -0.997 & -0.698 & -0.515 \\
LLM & -1.905 & -1.096 & -0.785 & -0.637 \\
\end{tabular}
\end{center}
\end{table}

\begin{table}
\caption{-hapax+lemmatized (zipf-mandelbrot parameters over 1-grams)}
\begin{center}
\begin{tabular}{r||rrr}
 & k & $\alpha$ & $\beta$ \\
\midrule\midrule
Education & 6444.49 & 1.92 & 1.11 \\
Inner Life & 11429.33 & 2.69 & 1.23 \\
Human & 1759.99 & -1.00 & 0.34 \\
LLM & 1521.47 & -1.00 & 0.34 \\
\end{tabular}
\end{center}
\end{table}

\begin{table}
\caption{-hapax-lemmatized (slope)}
\begin{center}
\begin{tabular}{r||rrrr}
 & 1-grams & 2-grams & 3-grams & 4-grams \\
\midrule\midrule
Education & -1.001 & -0.595 & -0.328 & -0.193 \\
Inner Life & -1.184 & -0.750 & -0.522 & -0.391 \\
Human & -1.617 & -0.954 & -0.675 & -0.497 \\
LLM & -1.810 & -1.051 & -0.764 & -0.624 \\
\end{tabular}
\end{center}
\end{table}

\begin{table}
\caption{-hapax-lemmatized (zipf-mandelbrot parameters over 1-grams)}
\begin{center}
\begin{tabular}{r||rrr}
 & k & $\alpha$ & $\beta$ \\
\midrule\midrule
Education & 3957.76 & 1.01 & 1.00 \\
Inner Life & 6024.59 & 1.43 & 1.08 \\
Human & 1691.61 & -1.00 & 0.35 \\
LLM & 1545.22 & -1.00 & 0.33 \\
\end{tabular}
\end{center}
\end{table}

As expected, for similar-sized corpora, the $\alpha$ and $\beta$
parameters are similar in all cases, indicating that the texts
do indeed follow a Zipfian distribution.
In all cases,
the LLM-generated corpora exhibit steeper slopes than
the comparable human-generated corpora,
and the slopes diverge more starkly at higher n-gram levels.
These observations validate hypothesis 1).
Lemmatization of the corpus had only negligible effects in all cases,
invalidating hypothesis 2).

The relationship can be seen even more clearly in the accompanying figures
\ref{fig:+hapax-lemmatizedloglinear}:

\begin{figure}[t]
  \centering

  \subfloat[+hapax-lemmatized 1-grams]{
      \includegraphics[width=.40\linewidth]{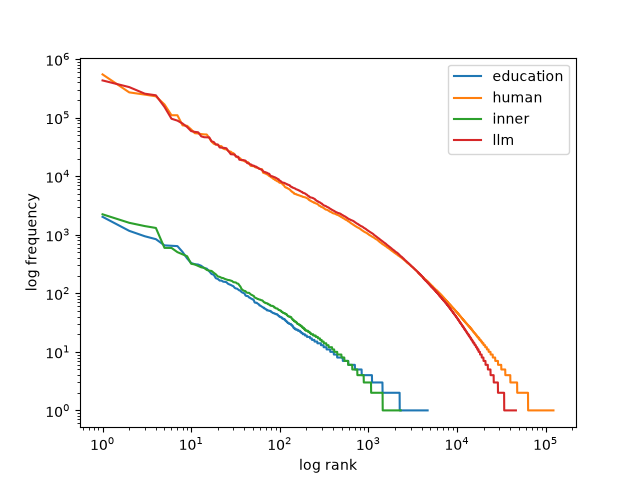}
  }\hfill

  \subfloat[+hapax-lemmatized 2-grams]{
      \includegraphics[width=.40\linewidth]{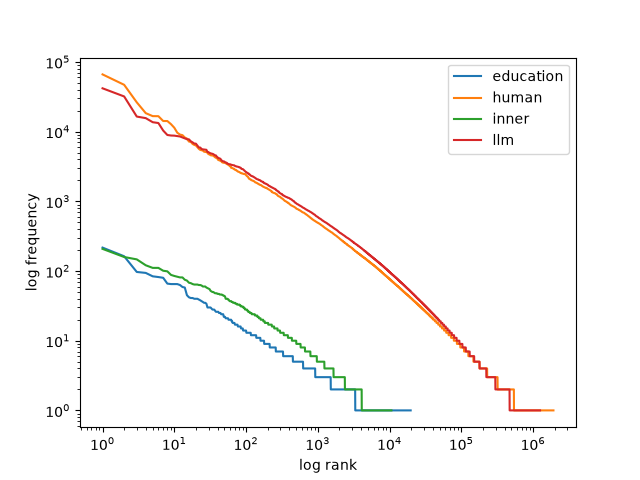}
  }\hfill

  \subfloat[+hapax-lemmatized 3-grams]{
      \includegraphics[width=.40\linewidth]{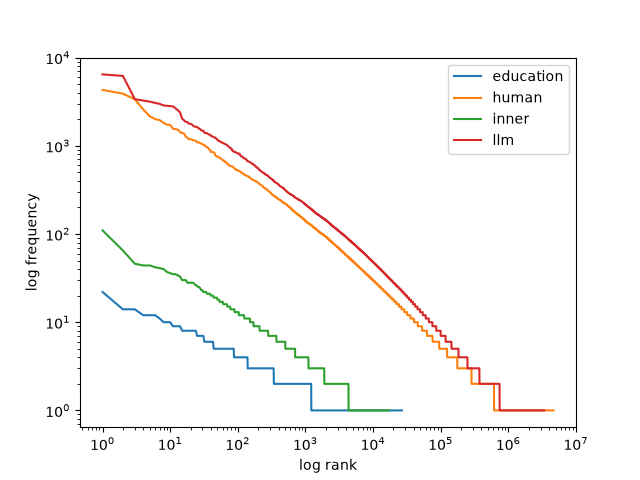}
  }\hfill

  \subfloat[+hapax-lemmatized 4-grams]{
      \includegraphics[width=.40\linewidth]{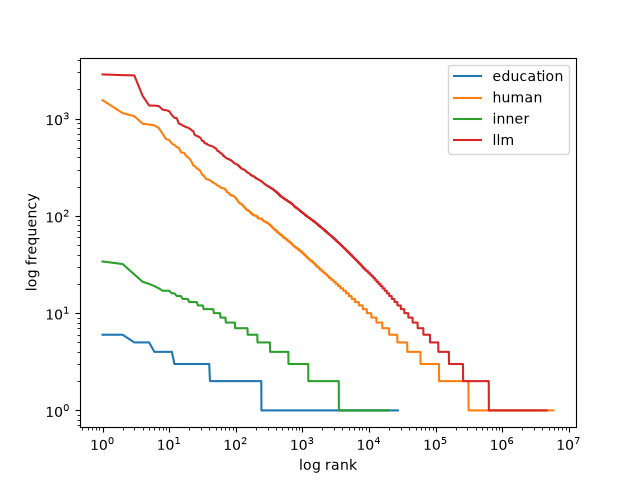}
  }\hfill

  \caption{Cross-corpus comparisons of log-log rank-frequency relationship at different n-gram levels, including hapax legomena}
  \label{fig:+hapax-lemmatizedloglinear}
\end{figure}

\begin{figure}[t]
  \centering

  \subfloat[-hapax-lemmatized 1-grams]{
      \includegraphics[width=.40\linewidth]{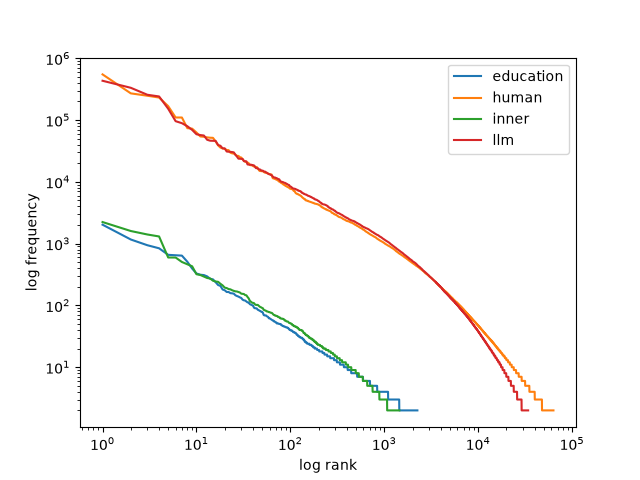}
  }\hfill

  \subfloat[-hapax-lemmatized 2-grams]{
      \includegraphics[width=.40\linewidth]{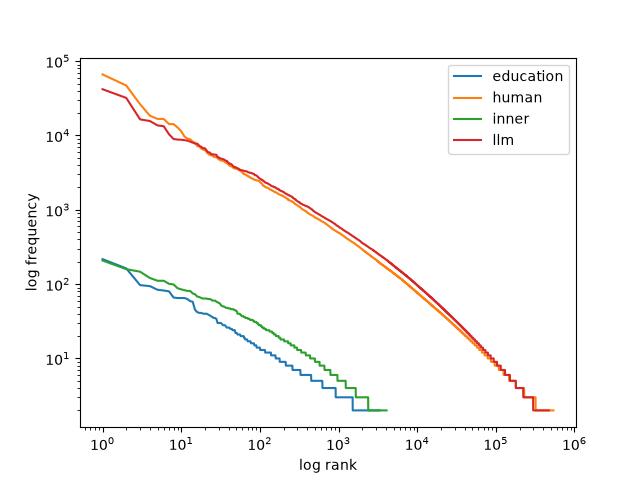}
  }\hfill

  \subfloat[-hapax-lemmatized 3-grams]{
      \includegraphics[width=.40\linewidth]{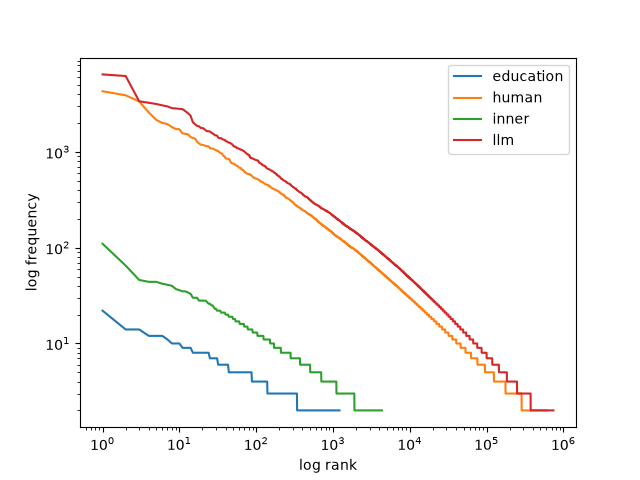}
  }\hfill

  \subfloat[-hapax-lemmatized 4-grams]{
      \includegraphics[width=.40\linewidth]{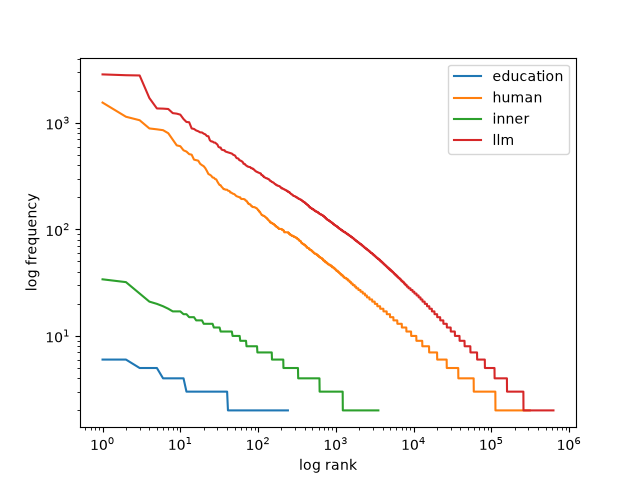}
  }\hfill

  \caption{Cross-corpus comparisons of log-log rank-frequency relationship at different n-gram levels, excluding hapax legomena}
  \label{fig:-hapax-lemmatizedloglinear}
\end{figure}

For 1-grams, the curves for LLM- and human-generated corpora look very similar.
At higher n-gram levels, the curves diverge noticeably:
more and more of the textual weight gravitates toward the most common
3- and 4-grams.
This tendency is apparent whether we include or exclude
\textit{hapax legomena}, indicating that these phenomena are not likely
to be the result of misspellings or one-off usages.

\section{Conclusions}

So far,
I have demonstrated a consistent distributional difference
between texts generated by LLMs and those written by humans.
Humans employ a larger vocabulary of both words and phrases
and demonstrate a much stronger preference for phrasal variety than LLMs.
These distributional differences affect LLMs' expressive range
in terms of both semantic content and stylistic/aesthetic palatability.

A direct consequence of the distributional characteristics of these corpora
is the following.
Consider a 4-gram language model built off of the human and LLM corpora.
For a given sequence of words \(w_1, w_2, w_3\),
the models can be used to generate \(w_4\).
Over multiple iterations,
the human-based model will generate more different \(w_4\) possibilities
than the LLM-based one.
If this model were then used iteratively to generate possibilities
for \(w_5\) and so on,
the human-based model would generate more sentences.
If one were to generate an equal number of sentences using each model,
the human-based model would generate more sentences.
As a coarse assumption,
identical word sequences encode identical semantic claims,
while divergent word sequenes encode different semantic claims.
Thus, the set of human-generated texts
would contain a greater amount of semantic content than the LLM-generated one.
In other words,
if given a fixed word limit within to generate text,
a human is likely to express more ideas than an LLM.

Literary style is harder to quantify.
Often, good literary style demands varied word choice and syntax,
but this is not always the case:
a well-deployed anaphora or other repetitive figure of speech
can lend rhetorical weight.
If LLMs managed variety and repetition in the same way that humans do,
we would expect to see similar distributional characteristics;
this is not the case,
so LLMs do somehow depart from human sensibilities
with respect to compositional use or avoidance of repetition.
It is beyond the scope of this contribution to diagnose
\textquote[ {\cite[2]{llmsdistort}} ]{the “feel” of LLM prose},
but the numbers contain some clues.
Anecdotally,
I find that LLMs rely on the reuse \textit{ad nauseam}
of preferred transitional phrases.
In \textit{The Inner Life of an AI},
the collocation "as an AI" occurs 110 times,
almost always deployed transitionally at the beginning of a sentence.
In \textit{The Education of Henry Adams},
the most-frequent trigram is "he could not,"
occurring only 22 times.
In the larger corpora,
one of the most frequent 4-grams is, ironically, "one of the most."
The LLMs generated this phrase 2787 times;
it shows up only 870 times in the large human-generated corpus.

In LLM-generated text,
low information density and poor stylistic "feel"
are symptoms of a common affliction:
lack of phrasal variety.
Style and meaning are connected.
If LLMs cannot compose aesthetically coherent prose,
then their ability to express will remain ineluctably impaired.

\bibliography{reference_main}

\end{document}